# Polygon-mamba: Retinal vessel segmentation using polygon scanning mamba and space-frequency collaborative attention

Yuanyuan Peng[a,*], Wen Li[a]

[a] School of Electrical and Automation Engineering, East China Jiaotong University, Nanchang, 330000, China

**Abstract**

Retinal vessel segmentation is crucial for diagnosis and assessment of ocular diseases. Notably, segmentation of small retinal vessels has been consistently recognized as a chal- lenging and complex task. To tackle this challenge, we design a hybrid CNN-Mamba fu- sion network that integrates polygon scanning mamba and space-frequency collaborative attention mechanism for the detection of small vessels. Considering that the traditional mamba architecture with horizontal- vertical scanning may compromise the topological integrity of target structures and result in local discontinuities in small retinal vessels, we present a polygon scanning visual state space model (PS-VSS) to identify small vessel structural features by multi-layer reverse scanning way. Which effectively preserves pixel connectivity, thereby substantially mitigating the loss of information pertaining to small vessels. Furthermore, as we all known that the spatial domain prioritizes positional and structural information, while the frequency domain emphasizes global perception and local detail components, a space-frequency collaborative attention mechanism (SFCAM) is introduced within the skip connection to extract efficient features from the spatial and frequency domains. This strategy empowers the model to dynamically enhance the key features while effectively suppressing clutters. To assess the efficacy of our model, it was tested on three publicly available datasets: DRIVE, STARE, and CHASE_DB1. Compared to manual annotations, our model demonstrated F1 scores of 0.8288, 0.8282, and 0.8251, Area Under Curve (AUC) values of 0.9808, 0.9840, and 0.9866, and Sensi- tivity (SE) values of 0.8356, 0.8314, and 0.8484 across three datasets, respectively. The effectiveness of our model was validated through both visual inspection and quantitative analysis.

## 1. Introduction

Segmentation of retinal vessels poses significant challenges due to various factors [1, 2]. Primarily, the weak contrast between retinal vessels and their surroundings, making them difficult to differentiate. Additionally, retinal vessels have a very complicated topology, with significant variations in scale and intricate interwoven patterns. This distinctive topology and image composition create obstacles for accurately segmenting retinal vessels. In clinical settings, manual segmentation of retinal vessels is particularly challenging [3].

* Corresponding author.
**E-mail addresses:** 8593@ecjtu.edu.cn **(Y. Peng)**

On the one hand, it requires a considerable amount of time. On the other hand, the subjective willingness and experience of different doctors lead to different segmentation results. Consequently, the application of deep learning technologies to aid clinicians in delineating the contours of retinal vessels has become an essential approach [4].

Numerous models and techniques have been developed to segment retinal vessels and yielding remarkable outcomes. In which, the UNet-based methods are renowned for their simplicity and exceptional performance [5]. However, they encounter challenges in managing long distance dependencies. Consequently, many researchers have devel- oped Transformer-based models that utilize self-attention mechanisms to achieve global parallel computing, effectively capturing long-distance dependencies [6]. Despite their ef- fectiveness, these techniques are often time-intensive. Recently, Mamba-based approaches have been introduced to mitigate this issue [7]. The primary advantage of Mamba-based methods is their linear time complexity, which allow for more efficient management of long-distance dependencies compared to traditional Transformer models. Despite nu- merous methods have been presented for detecting retinal vessels, there are still many limitations:

(I) Due to various factors such as lighting from different perspectives, complicated morphological topology, weak contrast between retinal vessels and their surroundings, and noise, it is difficult to segment microvessels.

(II) Many visual Mamba networks implement various scanning directions, such as parallel and diagonal. These intersecting scans often fail to maintain continuity along straight or diagonal lines, thereby reducing their sensitivity to complex structural direc- tions.

(III) Retinal vessels are intricately organized with complex structures, and relying solely on spatial attention information may lead to missing some key information.

To effectively tackle these challenges, we introduce a hybrid CNN-Mamba fusion net- work using a polygon scanning visual state space model (PS-VSS) and a space-frequency collaborative attention mechanism (SFCAM). Since conventional Mamba models adopt- ing horizontal and vertical scanning patterns tend to disrupt the topological integrity of anatomical structures and trigger fragmented continuity in tiny retinal capillaries, we pro- poses a PS-VSS module that leverages multi-layer reversed scanning strategy to capture fine structural patterns of microvessels. This approach preserves pixel-level connectiv- ity and significantly mitigates feature degradation in slender retinal vessels. Further- more, spatial representations emphasize contextual location and structural cues, while frequency-domain characteristics are adept at modeling long-range global correlations and intricate local details. Based on this understanding, we integrate a SFCAM into the skip connections. This multi-branch design thoroughly explores complementary repre- sentations across spatial and frequency dimensions, enabling the network to adaptively enhance prominent vessel features while suppressing extraneous interference. Our contri- butions are as follows:

(I) A novel segmentation model, Polygon-mamba, is presented for automatic and accu- rate detection of small retinal vessels, which substantially boosts segmentation accuracy.

(II) A PS-VSS module is introduced to efficiently extract semantic information by using a unique polygon scanning technology, thereby effectively capturing global features of retinal vessels and detecting small vessels.

(III) A SFCAM module is incorporated within the skip connection to extract effi- cient features from the spatial and frequency domains, allowing the model to adaptively highlight the feature representation of the target object while suppress clutters.

(IV) The PS-VSS and SFCAM modules are integrated into the UNet model to gen- erate an efficient retinal vessel segmentation scheme.

This paper is laid out as follows. In Section 2, we reviews numerous deep learning- based segmentation approaches. Section 3 gives a comprehensive explanation of the Polygon-mamba model. And Section 4 reveals experimental outcomes. Ultimately, the findings are discussed in Section 5, and concluded in Section 6.

## 2. Related work

### 2.1. CNN- based model

Traditional methodologies for retinal vessel segmentation have primarily focused on crafting robust features that align with spectral information and local image textures. Al- though high-resolution datasets offer precise geometric details and intricate textures, lim- ited feature extraction cannot effectively achieve efficient segmentation of retinal vessels. Fortunately, deep learning techniques have gained prominence in retinal vessel detection. Particularly, Fully Convolutional Networks (FCNs) have been employed to tackle seman- tic segmentation challenges [8], but their performance is often constrained. Recently, the UNet model has emerged as a popular choice in biomedical signal processing. It's encoder component reduces spatial dimensions and increases channel numbers, thereby enhanc- ing feature extraction efficiency. Concurrently, the decoder component upsamples the feature maps. Significantly, the skip connections in the UNet framework notably retain spatial and channel details, facilitating the incorporation of local and global information, which renders the UNet model particularly suitable for retinal image segmentation. Con- sequently, several improved versions of the UNet model have been developed, including Attention UNet [9], UNet++[10], LIOT[11], Swin-UNet[12], and Vm-UNet[13]. For in- stance, Shi et al. introduced the LIOT filter to maintain the structural characteristics of retinal vessels while minimizing clutters, subsequently achieving effective segmentation of these vessels using the IterNet model [11]. Nonetheless, this approach relies solely on intensity information, potentially resulting in weak vessels cannot be detected. To address this limitation, its improved version incorporates the ODoS filter to extract both local intensity and vector information, thereby improving the detection of weak vessels [14]. In contrast, Qi et al. employed a different strategy by integrating dynamic snake convolution with the UNet model to achieve effective vessel segmentation [15]. Similarly, Ma et al. developed the VasCA-Net, which adeptly extracts vascular features of varying widths through the design of EConv, DConv, and multi-scale channel attention modules [16]. Despite the effectiveness of these methods in extracting local features of the tar- get object, they exhibit limitations in global feature extraction, leading to suboptimal segmentation performance of retinal vessels.

### 2.2. Mamba- based model

CNNs are proficient in grasping local details, whereas Transformers are good at ex- tracting global semantics and long-range dependencies, these two methodologies naturally complement each other in feature extraction. However, Transformers are characterized by high computational complexity. Recently, Mamba has been introduced to facilitate global feature aggregation through a structured state space recursive mechanism, which effi- ciently captures long-distance continuous dependencies of retinal vessels while effectively managing computational and memory overhead. Based on this theory, extensive research has been undertaken on the application of Mamba in retinal vessel segmentation, result- ing in a series of improvement schemes have been presented, Such as vision mamba[17], Mamba-UNet [18], Serp-mamba [19], $S^3$-mamba [20], Mamba-sea [21], U-shape mamba [22] and GLMamba [23]. For example, Sun et al. developed a frequency-spatial entan- glement Mamba to highlight retinal vessel representation [24]. Nonetheless, this method exhibits suboptimal performance in segmenting small vessels. The limitation primar- ily stems from the inherent causal directional deviation defects present in the original Mamba architecture when applied to two-dimensional image segmentation. To overcome the question, Hu et al. introduced a Zigzag Mamba architecture, which enhances the network's positional awareness by employing a Zigzag scan path [25]. To further increase the effectiveness of segmentation, Liu et al. replaced the Zigzag approach with dynamic snake convolution, thereby enhancing the model's capability to identify complex vascular structures [26]. These methods improve the scanning style of the original mamba archi- tecture, forcing deep learning model to better recognize local information and complex retinal structural features in fundus images.

2.3. Attention mechanism

Numerous attention mechanisms have been developed and integrated into deep learn- ing models, enabling these approaches to focus on extracting features from retinal vessels. Building on this concept, various spatial attention mechanisms have been designed to emphasize key regions by adaptively assigning weights to different spatial positions[27]. Such as RCAR-UNet [28], CCS-UNet [29], MambaFuse [30] and MSFE-Mamba [31]. For instance, Varma et al. highlight retinal vessel representation by combining a modified Frangi filter with the classic attention UNet [32]. Similarly, Gao et al. embedded a novel rough attention within the GTSU-Net to accurately detect retinal vessels while reduc- ing computational complexity [33]. Using a different strategy, several multi-scale feature modules have been developed to address the unique inhomogeneous thickness structures of retinal vessels. For example, Bhimavarapu et al. [34] , Li et al. [35] , and Panchal et al. [36] extracted features of varying thicknesses of retinal vessels at different scales within the same level, focusing on diverse characteristics like textures, edges, and shapes of retinal vessels. In a similar vein, Yuan et al. [37], Li et al. [38], Luo et al. [39], and Cao et al. [40] have integrated shallow high-resolution features with deep semantic features, highlighting the complementary nature of information across different levels. While these methodologies have demonstrated efficacy in segmenting retinal vessels, they often over-look the critical role of frequency attention. To address this gap, Ma et al. developed the MSFB-Net, which incorporates multi-scale information in frequency domain and cross- layer boundaries specifically for image segmentation [41]. Furthermore, Yin et al. [42] and Li et al. [43] enhanced the segmentation performance of medical images by employing a wavelet transformer to establish frequency attention. Recently, Zhang et al. proposed an innovative approach for medical image segmentation by amalgamating directional, frequency-spatial, and structural attentions [44]. By integrating spatial and frequency attention, the efficiency of retinal vessel segmentation can be markedly improved.

Inspired by the aforementioned papers, the designed polygon scanning visual state space model and space-frequency collaborative attention mechanism are integrated into UNet model to generate an efficient retinal vessel segmentation scheme. Unlike other models, the proposed model focuses on segmenting small vessels, maintaining vascular continuity through the design of a polygon mamba, and integrating spatial attention and frequency attention to extract complex features of retinal vessels.

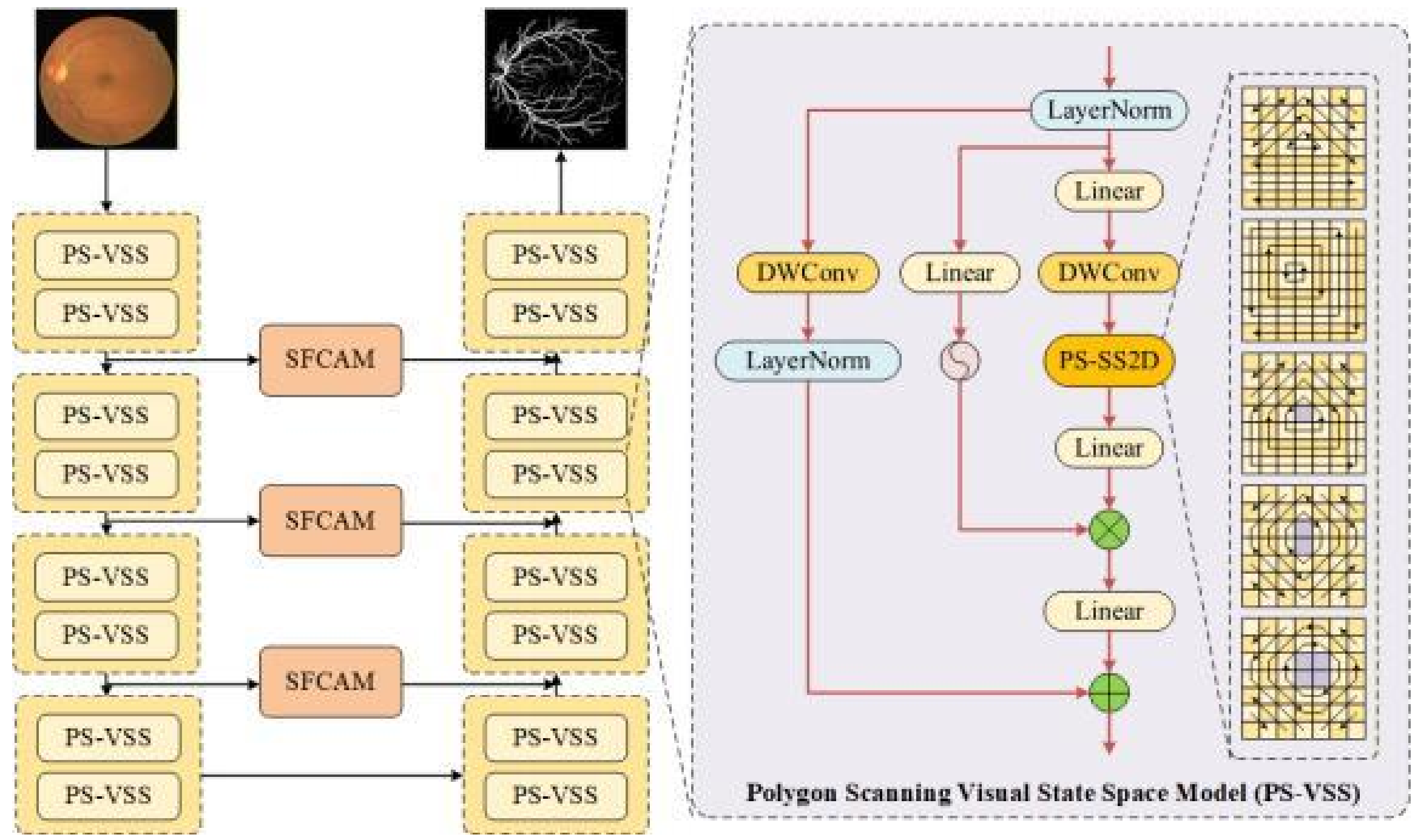

Figure 1. The presented model.

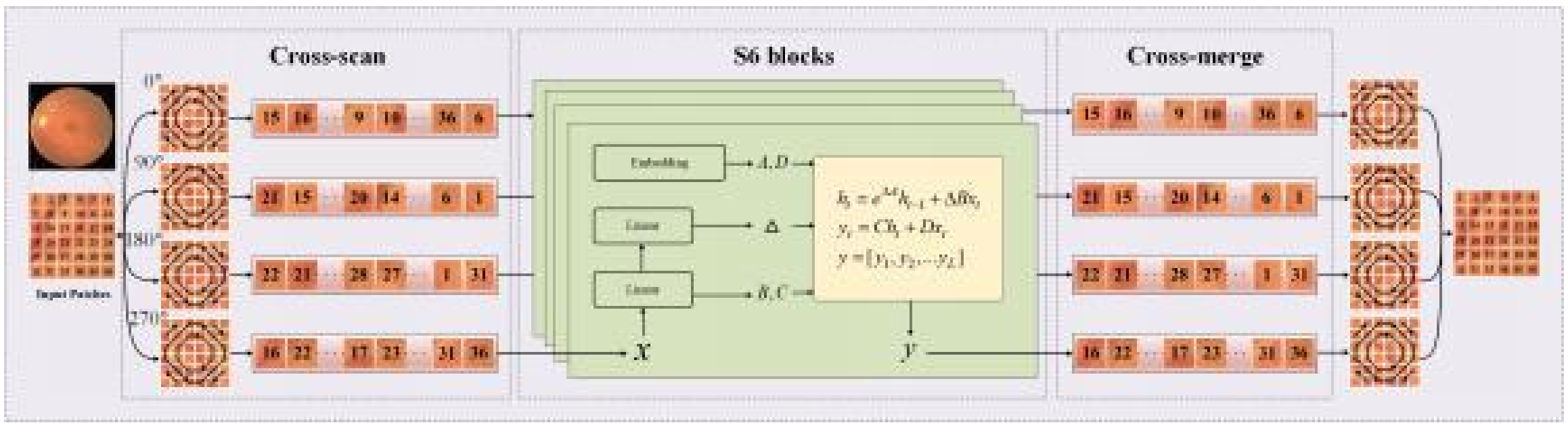

Figure 2. The improved visual state space model.

## 3. Method

### 3.1. Overview of the presented model

To achieve precise segmentation of microvessels, a hybrid CNN-Mamba fusion net- work based on polygon scanning mamba and space-frequency collaborative attention is designed, as depicted in Fig. 1. Initially, the PS-VSS module is intricately integrated with the CNN model to improve the capacity of feature maps, rendering it especially useful in managing complex background targets. Subsequently, a SFCAM module is de- veloped to dynamically and accurately capture the primary features of target objects, significantly enhancing the model's capacity to perceive feature maps and suppress clut- ters by integrating important information from both spatial and frequency domains. The strategy significantly reduces the loss of critical information while efficiently identifying small vessels.

### 3.2. Polygon scanning visual state space model (PS- VSS model)

Traditional state space model (SSM, S4) is primarily designed to handle one-dimensional sequences, with its dynamic parameters being predicted from the input sequence via a projection network. To extend the applicability of the SSM to two-dimensional spaces (SS2D), Liu et al. introduced VMamba[45], which proposes an input-dependent selection mechanism (SSM, S6) across multiple directions. As illustrated in Fig. 2, the SS2D frame- work is made up of three key parts: Cross-scan, S6 blocks, and Cross-merge. In which, the Cross-scan component transforms two-dimensional spatial features into multiple one- dimensional sequences oriented in various directions, enabling the one-dimensional state space model to process two-dimensional structures. The S6 block is responsible for per- forming selective state space modeling on each sequence, thereby capturing long-range dependencies with linear complexity and adaptively weighting relevant information. The Cross-merge component reconstructs the processed one-dimensional sequences back into two-dimensional spatial features and integrates multi-directional information to generate a compact and spatially coherent representation. Despite its widespread application in medical image processing, this method may not adequately fulfill the segmentation re- quirements for the detection of microvessels. The primary limitation of the traditional SS2D method lies in its reliance on horizontal and vertical scanning during the Cross-scan stage, which can compromise the integrity of the target topology and hinder the effective detection of small vessels.

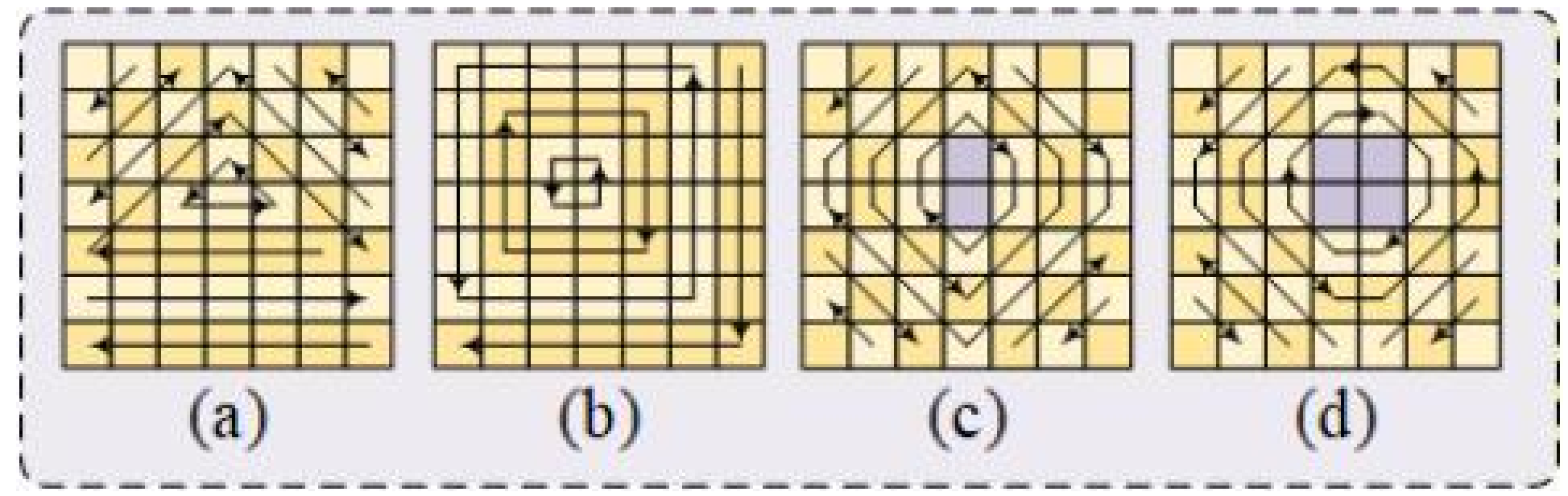

Figure 3. Polygon scanning SS2D way. (a)Triangular scan mode. (b)Quadrilateral scan method. (c)Hexagonal scan way. (d)Octagonal scan algorithm.

In order to effectively detect small vessels, we propose an innovative polygon scan- ning SS2D way. Unlike the conventional horizontal and vertical scanning techniques employed in the Cross-scan stage, our approach utilizes a serialized modeling process that initiates from the center and extends outward along a polygonal path, as illustrated in Fig. 3. Specifically, beginning with the determination of scanning origin points, a multi-directional pentagonal traversal sequence is generated by constructing four sets of scanning indices and their directional transformations, including $0^{o}$, $90^{o}$, $180^{o}$, and $270^{o}$ flipping paths. This approach facilitates a more comprehensive spatial rearrangement during the Cross-scan stage.

The multi-layer reverse scanning way systematically expands in a hierarchical manner from the innermost to the outermost layer, with the scanning direction of the M layer being opposite to that of the M + 1 layer, as shown in Fig. 4. For the N polygon, given the center point (a,b), the radius R, and the angle θ between a specific vertex and the X-axis, the coordinates $(A_i, B_i)$ of all vertices can be described as:

$$A_i = a + R\cos\left(\theta + \frac{2\pi}{N}i\right), \quad i = 0, 1, \ldots, N-1 \tag{1}$$

$$B_i = b + R\sin\left(\theta + \frac{2\pi}{N}i\right), \quad i = 0, 1, \ldots, N-1 \tag{2}$$

Once all vertices are identified, adjacent vertices are connected to form the polygon.

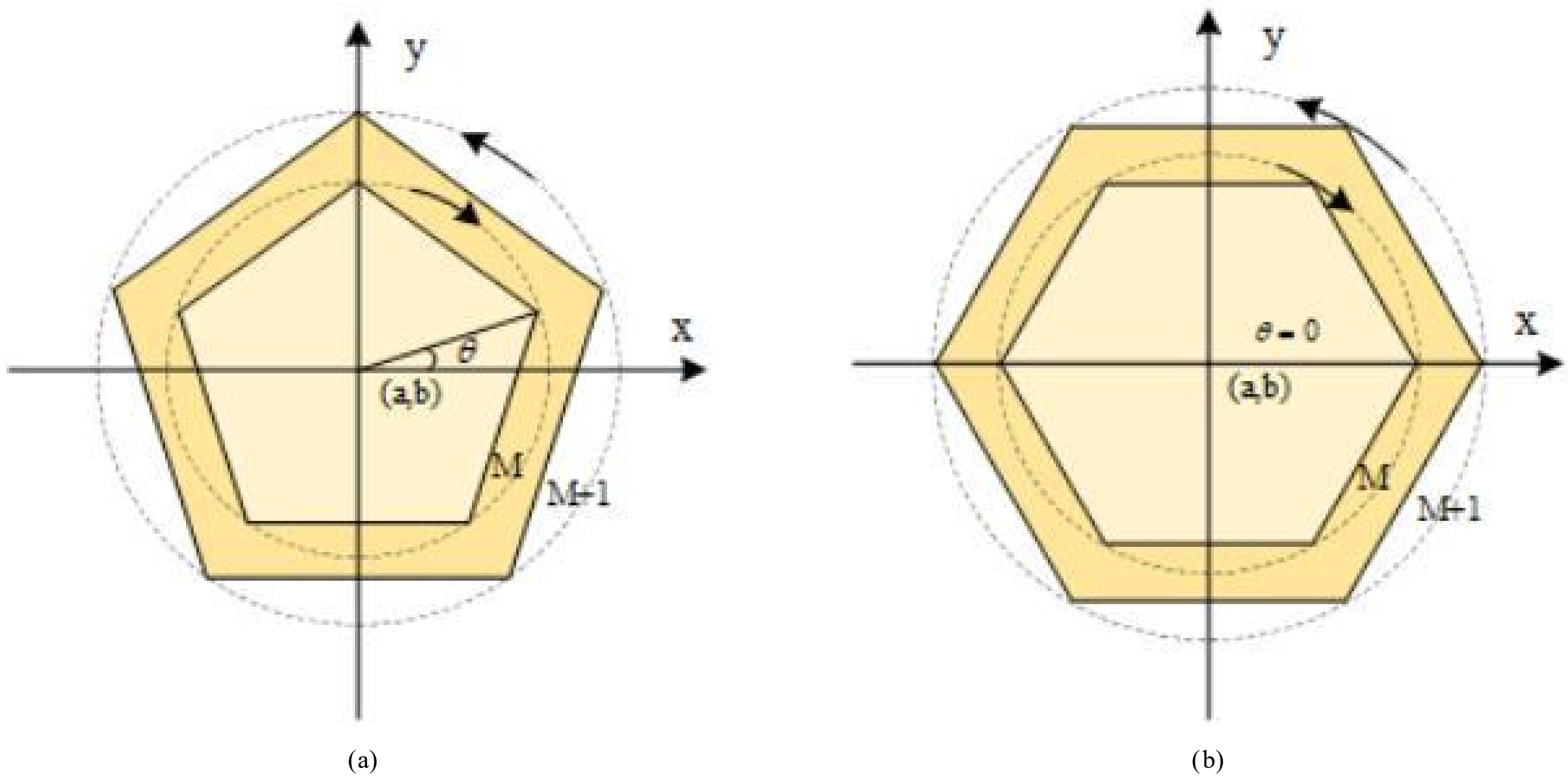

Figure 4. Multi-layer reverse scanning way. (a)Pentagon. (b)Hexagon.

all spatial positions of the polygon that meet the specified distance constraints are ini- tially obtained, and then sorted based on their angular order relative to the center point to establish the traversal sequence for the special layer. Specifically, forward scanning begins from the position with the smallest angle, while reverse scanning starts from the position with the largest angle. Thus, the entire two-dimensinal feature map is organized into a polygonal scanning sequence that progressively expands outward layer by layer from a central starting point. The core of this design lies in enhancing the state space unit's ability to model slender, curved, and branching vascular structures through the implementation of diversified scanning paths, thereby improving the representation of complex vascular topology and spatial continuity. Compared to the original Mamba, the designed model not only retains the advantages in global structural modeling, but also preserves vascular topology continuity, altering it to enhance its suitability for vascular continuity, branching topology, and boundary precision in retinal vessel segmentation tasks.

### 3.3. Space-frequency collaborative attention mechanism (SFCAM)

It is well established that the spatial domain prioritizes positional and structural information, while the frequency domain emphasizes global perception and local detail components. To improve the segmentation performance, we propose a space-frequency collaborative attention mechanism (SFCAM) to effectively extract global features and detailed information of retinal vessels. As illustrated in Fig. 5, the SFCAM module comprises three branches for feature mapping. Here, the local branch is designed for extracting features at various scales, the global branch used the polygon scanning mamba for global feature extraction, and the Wavelet Transformer branch (WT) is applied to transform original features into high-frequency and low-frequency components. Then, we design a bidirectional cross-attention fusion module (BCFM) that combines local with high-frequency information, and global with low-frequency information to achieve stronger mixed features. Ultimately, the global, local, and mixed features are integrated to segment retinal vessels. This strategy can effectively solve the challange of insufficient exploitation of complex topological structures and details in the transmission process of complex features, thereby highlighting the representation ability of microvessels, edge contours, and local details while maintaining vascular topological continuity.

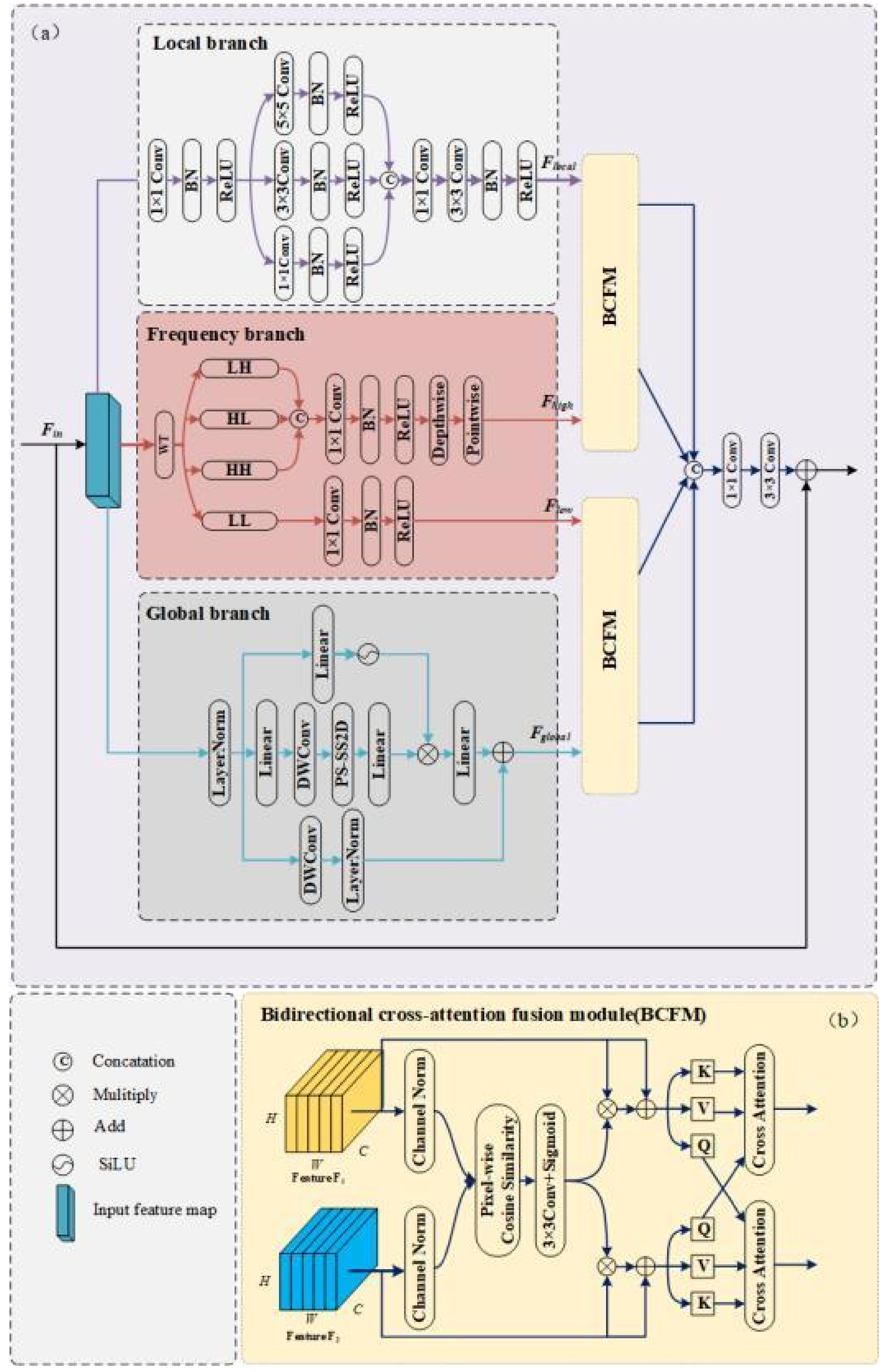

Figure 5. The Space-frequency collaborative attention mechanism (SFCAM). (a)Space-frequency collab- orative attention mechanism complete module. (b)Bidirection cross-attention fusion module.

Particularly, for an input feature $F_{in} \in R^{C\times H\times W}$, where C , H and W are the channels, height and width of the input feature, respectively. Subsequently, $F_{in}$ is used as input to obtain local features $F_{local}$, global features $F_{global}$, and frequency features $F_{low}$ and $F_{high}$ through three branches:

$$F_{local} = f_l(F_{in}) \tag{3}$$

$$F_{global} = f_g(F_{in}) \tag{4}$$

$$[F_{low}, F_{high}] = f_w(F_{in}) \tag{5}$$

Where $f_l(\cdot), f_g(\cdot)$, and $f_w(\cdot)$ represent the local branch, global branch, and WT branch for feature association. Subsequently, the spatial features and frequency-domain features are divided into two distinct groups and processed through the BCFM module. This module serves to align the semantic discrepancies between frequency and spatial features, facilitating effective feature selection. Lastly, the selected features are merged for retinal vessel segmentation.

$$[F_{local}^{Cross}, F_{high}^{Cross}] = f_{bcfm}(F_{local}, F_{high}) \tag{6}$$

$$[F_{global}^{Cross}, F_{low}^{Cross}] = f_{bcfm}(F_{global}, F_{low}) \tag{7}$$

$$F_{cat} = (Cat(F_{local}^{Cross}, F_{high}^{Cross}, F_{global}^{Cross}, F_{low}^{Cross}))) + F_{in} \tag{8}$$

$$F_{out} = Conv_{3\times3}(Conv_{1\times1}(F_{cat}))) + F_{in} \tag{9}$$

Where $f_{bcfm}(\cdot)$ denotes the BCFM module for aligning two different features, and $F_{out}$ represents the mixed features.

1) In the local branch, the input F undergoes 1 × 1 convolution operation, followed by multi-scale processing (1 × 1, 3 × 3, 5 × 5) to obtain multi-scale information of retinal vessels. Finally, $F_{local}$ can be obtained through 1 × 1 and 3 × 3 convolution operations.

$$F_b = Conv_{1\times1}(F_{in}) \tag{10}$$

$$F_c = Cat(Conv_{1\times1}(F_b), Conv_{3\times3}(F_b), Conv_{5\times5}(F_b)) \tag{11}$$

$$F_{local} = Conv_{3\times3}(Conv_{1\times1}(F_c)) \tag{12}$$

The purpose is to concurrently capture multi-scale features of both coarse and fine retinal vessels.

2) Within the global branch, the input $F_{in}$ undergoes normalization and local spa- tial preprocessing. Subsequently, an adaptive scanning sequence is developed along the primary pathway, initiating from the central region and extending outward in accor- dance with the geometric form of the polygon. This approach facilitates the sequential recombination of two-dimensional features in various path orders, thereby achieving com- prehensive aggregation of global context and directional dependencies through SSM.

$$X_1 = LN(PS - SS2D(DWConv(LN(F_{in})))) \tag{13}$$

The local compensation path directly enhances the local spatial response of input features to highlight high-resolution structural responses such as small blood vessels, boundary contours, and local fracture areas.

$$X_2 = \text{SiLU}(\text{Linear}(\text{LN}(F_{in}))) \quad (14)$$

Finally, the outputs of the two paths are combined in the feature space and combined with residual connections to generate the final result.

$$F_{global} = X_1 \times X_2 + \text{DWConv}(\text{LN}(F_{in})) \quad (15)$$

The objective is to develop long-range dependencies and global contextual information within retinal images, thereby effectively preserving the connected topological structure of vessels.

3)In the WT branch, the input $F_{in}$ is initially decomposed into low-frequency and high-frequency sub signals ( $F_{LL}$ , $F_{LH}$ , $F_{HL}$ and $F_{HH}$ ) using Discrete Wavelet Transform. This decomposition enables the network to model and interact with different frequency components separately. Here, the low-frequency component primarily retains smoother information, such as the morphology of the main blood vessels, regional connectivity, and overall topology. In contrast, the high-frequency component predominantly captures detailed features, including blood vessel boundaries, small branches, and local texture variations.

$$[F_{LL}, F_{LH}, F_{HL}, F_{HH}] = \text{WT}(F_{in}) \quad (16)$$

$$F_H = \text{Cat}(F_{LH}, F_{HL}, F_{HH}) \quad (17)$$

In which, WT represents the two-dimensional discrete wavelet transform operator. Subsequently, the 1 × 1 convolution, Depthwise Convolution (DW), and Pointwise Convolution (PW) are combined to get the high-frequency feature $F_{high}$ .

$$F_{high} = \text{PW}(\text{Conv}(\text{DW}(\text{Conv}(F_H)))) \quad (18)$$

For the low-frequency component $F_{LL}$ , a 1 × 1 convolution is used for channel align- ment to align it with high-frequency features in dimensional space.

$$F_{low} = \text{Conv}_{1\times1}(F_{LL}) \quad (19)$$

The purpose is to selectively screen and enhance effective high and low frequency features, strengthen details and structural expression.

To adaptively fuse the above multi-branch features, we propose a BCFM module to extract more robust mixed features. Initially, channel normalization, pointwise, and sigmoid operations are used to the two input features $F_{local}$ and $F_{high}$ to produce spatially correlated maps that explicitly describe the degree of consistency of different features in spatial positions.

$$W_{lh} = \sigma(\text{Conv}(\prod_{c=1}^{c})\text{LN}(F_{local} \odot \text{LN}(F_{high}))) \quad (20)$$

$$F'_{local} = F_{local} \odot W_{lh} + F_{local} \quad (21)$$

$$F'_{high} = F_{high} \odot W_{lh} + F_{high} \quad (22)$$

Among them, LN is the explicitly normalization, $\odot$ represents element multiplica- tion, Conv denotes the related refinement function composed of convolution operations, $\sigma$ represents the Sigmoid activation function, $\prod_{c=1}^{c}$ represents element wise multiplica- tion on the Cth channel. Using the same strategy, the correlated maps $F'_{low}$ and $F'_{global}$ can be obtained. To avoid the positional shift caused by one-way information transmis- sion, a query exchange strategy is adopted, which constructs bidirectional cross attention between the two branches separately, so that different features can be selected and re- selected simultaneously in the symmetrical information flow, thus fully exploring their complementarity. By using the cross multihead attention, a bidirectional cross attention is constructed to dynamically reassign features by integrating the query of one branch's features with the key and value of another branch's features, achieving bidirectional in- teraction, concatenation, and fusion.

$$Q_l, K_l, V_l = \delta_{qkv}^{lh}(F'_{local}) \tag{23}$$

$$Q_h, K_h, V_h = \delta_{qkv}^{lh}(F'_{high}) \tag{24}$$

$$F_{low}^{Corss} = \text{SoftMax}\left(\frac{Q_h K_l^T}{\sqrt{d_k}}\right) V_l \tag{25}$$

$$F_{high}^{Corss} = \text{SoftMax}\left(\frac{Q_l K_h^T}{\sqrt{d_k}}\right) V_h \tag{26}$$

Among them, $d_k$ represents the scaling factor, $\delta$ represents a learnable linear transfor- mation. Using the same strategy, the cross attention $F_{global}^{Cross}$ and $F_{low}^{Cross}$ can be obtained. Finally, all of the output features are concatenated in the channel dimension, and fea- ture fusion and reconstruction are performed through convolution operations to obtain a unified output representation, which is then connected using residual connections. This design allows the model to integrate fine details with global structural information while maintaining feature diversity.

$$F_{cat} = (\text{Cat}(F_{local}^{Cross}, F_{high}^{Cross}, F_{global}^{Cross}, F_{low}^{Cross}))) + F_{in} \tag{27}$$

$$F_{out} = \text{Conv}_{3\times3}(\text{Conv}_{1\times1}(F_{cat}))) + F_{in} \tag{28}$$

The SFCAM module deeply integrates frequency information and spatial information, effectively extracting local and global features of retinal vessels, leading to more precise and detailed segmentation of small vessels.

## 4. Experimental Results

### 4.1. Experimental Conditions and Datasets

Table 1. Details of the three experimental datasets used in this study.

| Datasets | Quantity | Resolution | Train- Test Allocation |
| --- | --- | --- | --- |
| DRIVE | 40 | 565 × 584 | 20 –20 |
| STARE | 20 | 605 × 700 | 10 – 10 |
| CHASE_DB1 | 28 | 999 × 960 | 20 – 8 |

The experimental environment is configured as follows: the graphics card is GeForce RTX 4090, system memory is 24GB RAM with batch sizes 64 and 100 epochs. To evalu- ate the algorithm's capability in segmenting retinal vessels, experimental validation was conducted across three datasets, as shown in Table 1. Simultaneously, preprocessing operations were applied to these datasets to mitigate the issue of limited dataset size.

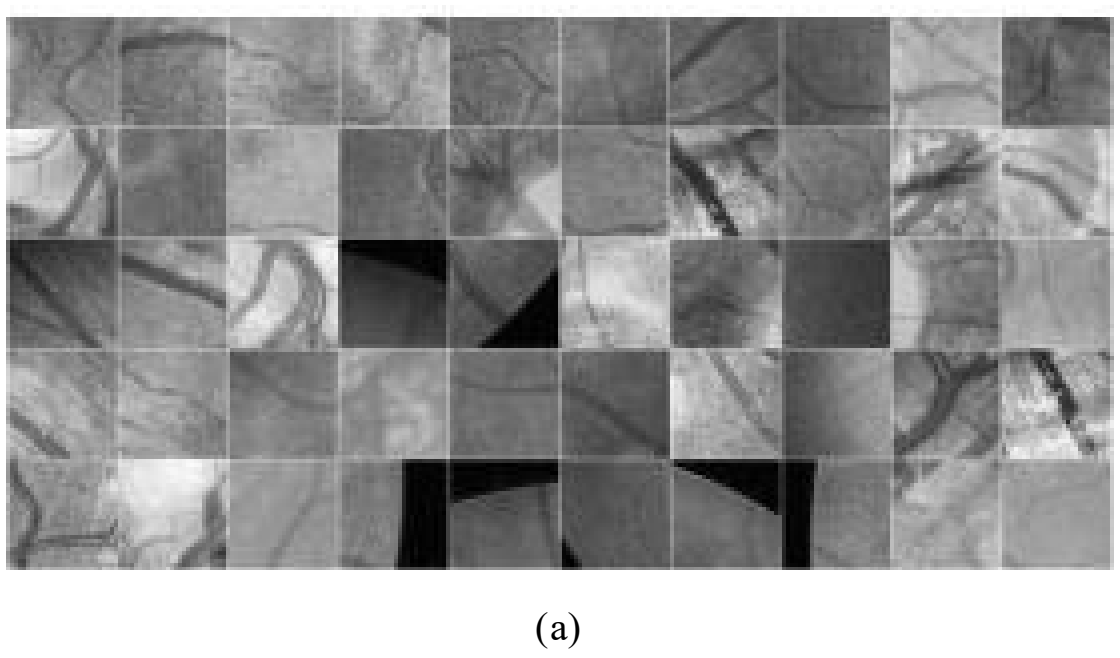

(a)

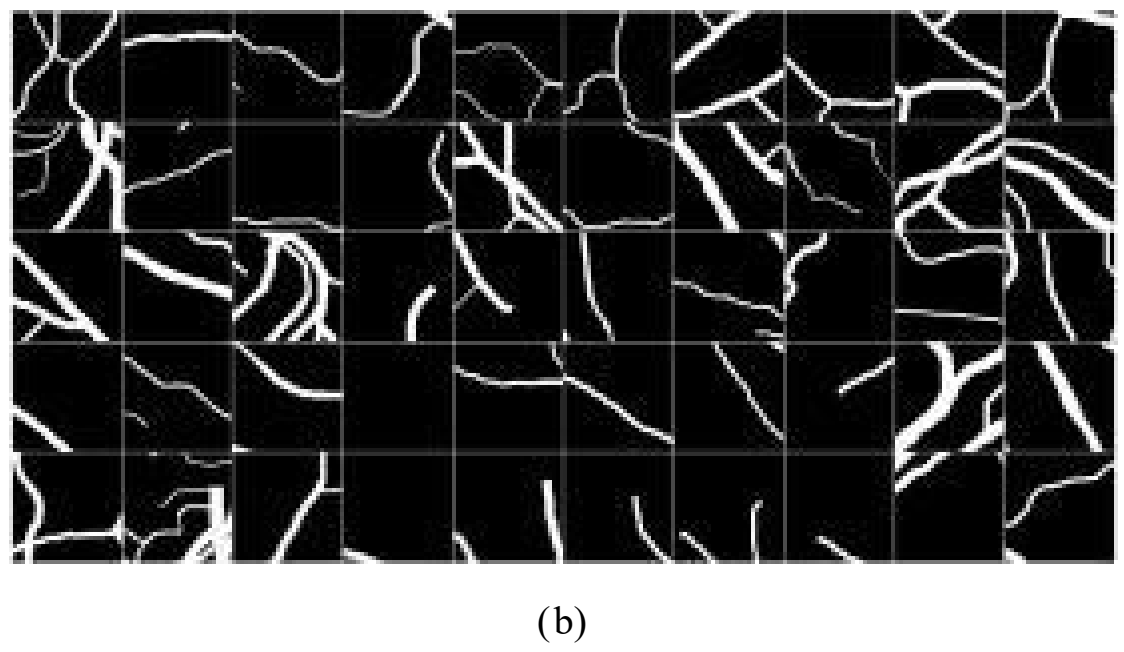

(b)

Figure 6. Pre-processing operations involving the division of three datasets into smaller chunks: (a) sample input images, and (b) sample input mask.

Specifically, each original image was partitioned into 16 smaller images, which were subsequently subjected to random rotation and rearrangement, with noise added to augment three datasets. An illustration of these datasets subdivision is depicted in Fig. 6.

4 .2. Evaluation Metrics

TP (True Positive): True positive indicates a sample that is genuinely positive and accurately identified as positive by the model.

FP (False Positive): False positive refers to a sample that is actually a negative class but is incorrectly predicted as a positive class by the model.

TN (True Negative): True negative refers to a sample that is actually negative and correctly predicted as negative by the model.

FN (False Negative): False negative refers to a sample that is actually positive but incorrectly predicted as negative by the model.

To demonstrate the effectiveness of our scheme, we use commonly used Intersection over Union (IOU), Area Under Curve (AUC), F1, Accuracy (ACC), Sensitivity (SE), and Specificity (SP) indicators to evaluate the validation in three publicly used datasets.

$$\mathrm{SE} = \frac{TP}{TP+FN} \tag{29}$$

$$\mathrm{SP} = \frac{TN}{TN+FP} \tag{30}$$

$$\mathrm{PR} = \frac{TP}{TP+FP} \tag{31}$$

$$\mathrm{IoU} = \frac{TP}{TP+FP+FN} \tag{32}$$

$$\mathrm{F1} = \frac{2\times PR\times SE}{PR+SE} \tag{33}$$

$$\mathrm{Acc} = \frac{\mathrm{TP} + \mathrm{TN}}{\mathrm{TP} + \mathrm{TN} + \mathrm{FP} + \mathrm{FN}} \tag{34}$$

Among these metrics, IoU measures the overlap between the predicted area and the actual annotated area, AUC is the core evaluation metric for assessing the quality of binary classification models, ACC refers to the rate of correct classification for both positive and negative cases across all samples, and F1 denotes the correlation between actual data and segmentation predictions. SE reflects the ability of a model to correctly detect all positive samples. while SP assesses the capability of a model to correctly distinguish all negative samples.

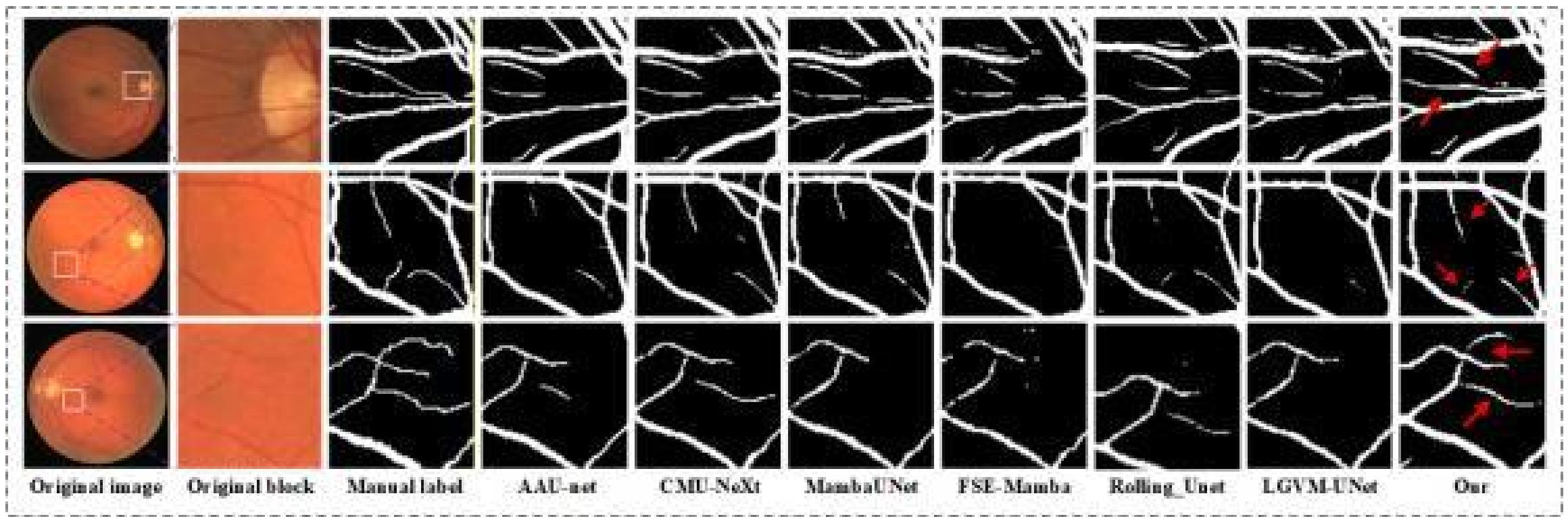

Figure 7. Retinal vessel segmentation on the DRIVE dataset

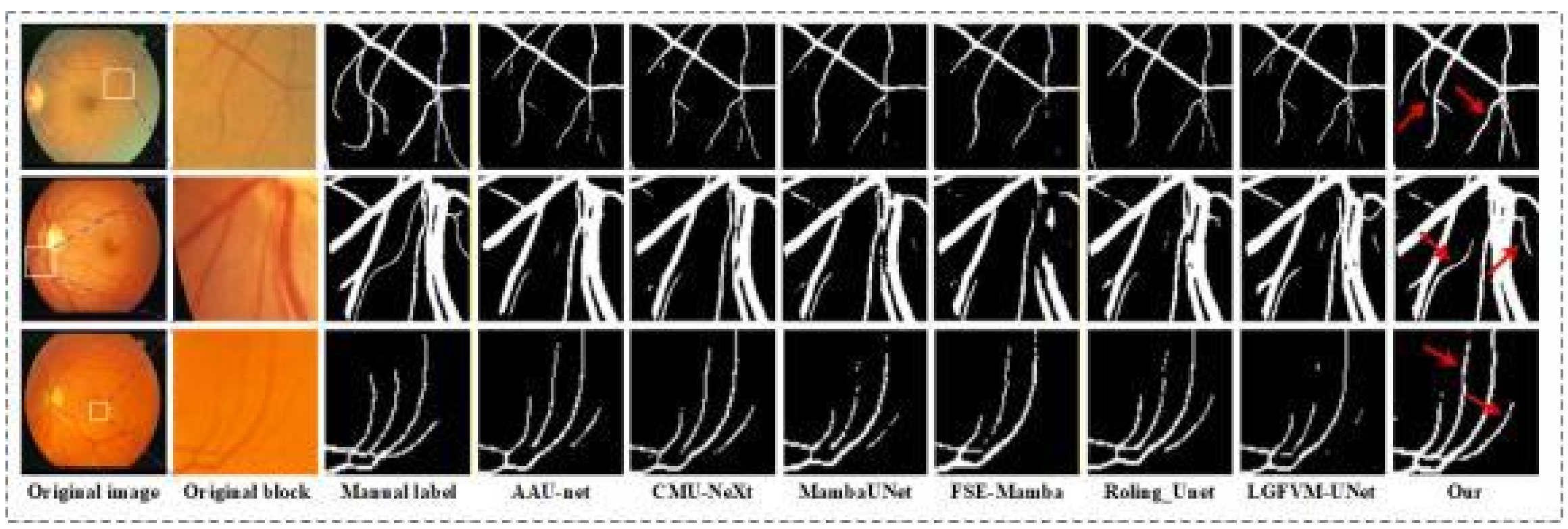

Figure 8. Retinal vessel segmentation on the STARE dataset

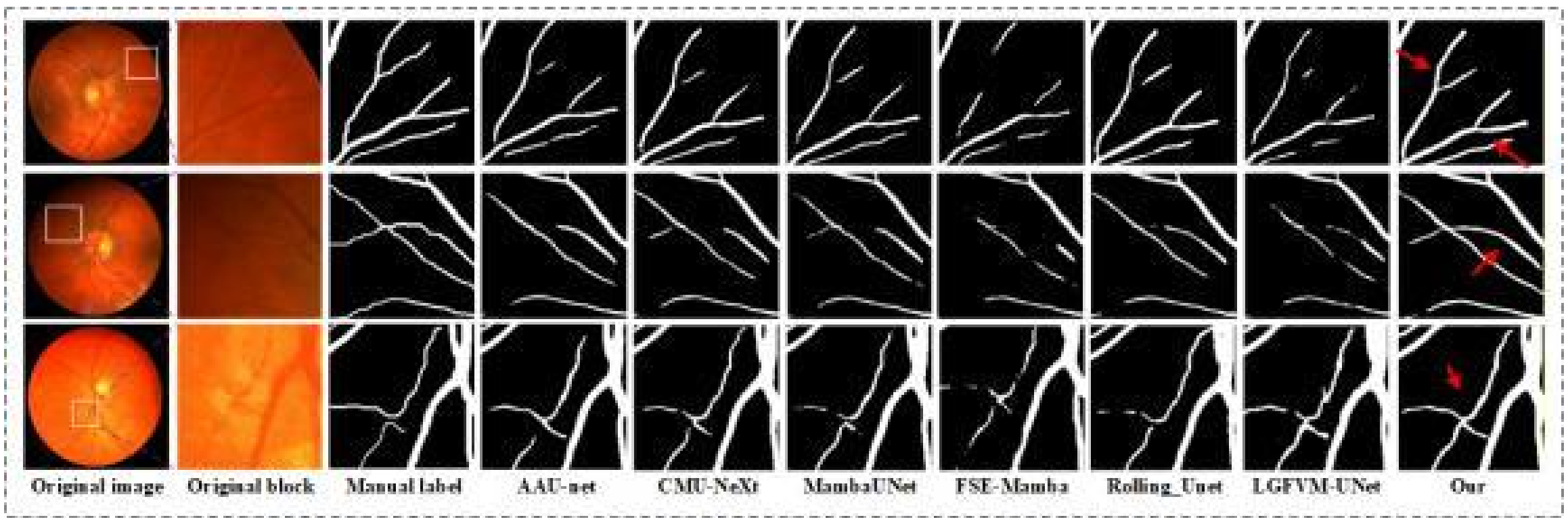

Figure 9. Retinal vessel segmentation on the CHASE_DB1 dataset

### 4 .3. Evaluate the presented method

#### 4 .3. 1 . Visual evaluation

To facilitate a more intuitive visual representation of the retinal vessel segmentation, Fig. 7, 8 and 9 offer a comprehensive visual comparison derived from some algorithms on the three datasets. The visualisations juxtapose our model with many state-of-the-art methods, including AAU-net [46], MambaU-Net[18], FSE-Mamba[24], Rolling_Unet [47], and LGFVM-UNet [48]. By focusing on specific regions, we present a validate assessment of the segmentation performance in fine vessels, thereby highlighting our algorithm's ability to accurately segment retinal vessels. On the DRIVE dataset, Polygon mamba effectively recovers fine vessels frequently overlooked by other models, while maintaining distinct vessel structures. On the STARE dataset, Polygon mamba exhibits enhanced robustness, effectively preserving minute branches and vessel crossings with minimal ar- tifacts. Sarly, on the CHASE_DB1 dataset, Polygon mamba achieves the best segmenta- tion performance on winding and thick vessels, closely resembling manual annotations. In contrast, the contrast models exhibit local discontinuities and missing fine vessels, indicating its limited ability to adapt to different types of vessels. The proposed Poly- gon mamba mitigates these limitations by incorporating CNN, PSS-VSS, and SFCAM modules, which improve the model's adaptability in both spatial and frequency domains.

4 .3.2. Quantitative evaluation

To assess the segmentation performance of the Polygon-mamba, we conducted a com- parative analysis against several methods. Quantitative results derived from the three datasets are presented in Tables 2, 3,and 4, with the highest scores emphasized in bold.

As illustrated in Table 2, the proposed method attained the F1 score of 0.8288, AUC value of 0.9808, and SE score of 0.8356, outperforming all comparative methods. This demonstrates that our scheme outperforms all comparative approaches in retinal vessel detection. Compared with the KM-Unet[53], the Polygon-mamba exhibited improve- ments of 0.22% and 2.94% in F1 and SE. respectively. Even though its ACC value (0.9561) was 0.08% lower than the KM-Unet[53]'s (0.9569), the difference in performance is minimal. The quantitative results from the DRIVE dataset further substantiate the superior ability of our scheme in detecting small retinal vessels.

On the STARE dataset, the performance comparison demonstrates that our model achieved the best F1 score (0.8282), AUC value (0.9840), and SE value (0.8314), as de- picted in Table 3. Compared with the Rolling-UNet[47], the Polygon-mamba exhibited improvements of 0.47%, 0.04% and 6.82% in F1, AUC and SE, respectively. These find- ing suggest that Polygon-mamba demonstrates strong generalization capability in the evaluation on the STARE dataset, effectively balancing fine-grained detection and back- ground. Although the ACC value of our model is lower than that of the Rolling-Unet[47], the SE value of our model is much higher than that of the comparison method. In other words, our model is significantly superior to the comparative methods in detecting retinal vessels.

Experimental results for the CHASE_DB1 dataset are illustrated in Table 4. Polygon-mamba achieved the highest SE (0.8484), AUC (0.9866), and F1(0.8251), significantly surpassing other methods. Although its SP (0.9770) was lower than FSE-Mamba[24] model (0.9897), the Polygon-mamba achieved the hight SE (0.8484), leading the ACC of our model (0.9642) was higher than FSE-Mamba[24] model (0.9597). In other words, our model achieved a more balanced SE and SP trade-off.

In summary, the presented method consistently exhibited robust performance across three datasets, achieving notable improvements in F1 and AUC. Illustrating its excellent adaptability and generalization in retinal vessel segmentation. Furthermore, the proposed method obtained the highest SE value, indicating superior performance in segmenting retinal vessels compared to other methods.

Despite these improvements, some limitations remain. The SP value of the presented method is slightly lower than some specific models, our model has unsatisfactory perfor- mance in noise suppression. The unsatisfactory performance in noise suppression is the main reason for its lower ACC value compared to other methods. Future work will focus on enhancing SP while maintaining SE to improve the segmentation of retinal vessels.

Table 2. Comparison of various models on the DRIVE dataset.

| Methods | Years | IOU | AUC | F1 | ACC | SE | SP |
|---|---|---|---|---|---|---|---|
| CS2-Net [49] | 2021 | 0.6964 | 0.9462 | 0.8210 | 0.9556 | 0.8001 | 0.9783 |
| AAU-net [46] | 2022 | 0.6982 | 0.9745 | 0.8223 | 0.9560 | 0.7991 | 0.9789 |
| CMU-NeXt [50] | 2024 | 0.7011 | 0.9793 | 0.8243 | 0.9567 | 0.7990 | 0.9797 |
| Mamba-UNet [18] | 2024 | 0.6900 | 0.9787 | 0.8166 | 0.9558 | 0.7729 | 0.9825 |
| LMFR-Net [51] | 2025 | 0.7007 | 0.9806 | 0.8240 | 0.9570 | 0.7904 | 0.9813 |
| FSE-Mamba [24] | 2025 | 0.6801 | 0.9769 | 0.8096 | 0.9547 | 0.7572 | **0.9835** |
| Rolling-Unet [47] | 2025 | 0 7024 | 0 9804 | 0 8252 | 0 9572 | 0 7933 | 0 9812 |
| MDF-Net [52] | 2026 | 0.7008 | 0.9727 | 0.8241 | 0.9550 | 0.8288 | 0.9734 |
| KM-UNet [53] | 2026 | 0.7044 | 0.9802 | 0.8266 | **0.9569** | 0.8062 | 0.9789 |
| LGFVM-UNet [48] | 2026 | 0.6713 | 0.9762 | 0.8083 | 0.9542 | 0.7590 | 0.9827 |
| Our | 2026 | **0.7076** | **0.9808** | **0.8288** | 0.9561 | **0.8356** | 0.9736 |

Table 3. Comparison of various models on the STARE dataset.

| Methods | Years | IOU | AUC | F1 | ACC | SE | SP |
|---|---|---|---|---|---|---|---|
| CS2-Net [49] | 2021 | 0.6853 | 0.9612 | 0.8133 | 0.9632 | 0.7548 | 0.9880 |
| AAU-net [46] | 2021 | 0.6892 | 0.9729 | 0.8160 | 0.9639 | 0.7557 | 0.9886 |
| CMU-NeXt [50] | 2022 | 0.6856 | 0.9797 | 0.8135 | 0.9631 | 0.7578 | 0.9875 |
| Mamba-UNet [18] | 2024 | 0.6715 | 0.9825 | 0.8035 | 0.9617 | 0.7384 | 0.9881 |
| LMFR-Net [51] | 2024 | 0.6887 | 0.9835 | 0.8157 | 0.9637 | 0.7566 | 0.9883 |
| FSE-Mamba [24] | 2025 | 0.6391 | 0.9803 | 0.7798 | 0.9585 | 0.6859 | **0.9908** |
| Rolling-Unet [47] | 2025 | 0 6999 | 0 9836 | 0 8235 | **0.9653** | 0.7632 | 0.9893 |
| MDF-Net [52] | 2026 | 0.6903 | 0.9672 | 0.8169 | 0.9629 | 0.7792 | 0.9848 |
| KM-UNet [53] | 2026 | 0.6883 | 0.9827 | 0.8154 | 0.9639 | 0.7522 | 0.9890 |
| LGFVM-UNet [48] | 2026 | 0.6574 | 0.9826 | 0.7933 | 0.9605 | 0.7137 | 0.9898 |
| Our | 2026 | **0.7069** | **0.9840** | **0.8282** | 0.9634 | **0.8314** | 0.9791 |

Table 4. Comparison of various models on the CHASE_DB1 dataset.

| Methods | Years | IOU | AUC | F1 | ACC | SE | SP |
|---|---|---|---|---|---|---|---|
| CS2-Net [49] | 2021 | 0.6852 | 0.9707 | 0.8132 | **0.9651** | 0.7657 | 0.9871 |
| AAU-net [46] | 2022 | 0.6734 | 0.9801 | 0.8048 | 0.9633 | 0.7615 | 0.9856 |
| CMU-NeXt [50] | 2024 | 0.6829 | 0.9860 | 0.8116 | 0.9647 | 0.7658 | 0.9866 |
| Mamba-UNet [18] | 2024 | 0.6788 | 0.9863 | 0.8087 | 0.9645 | 0.7542 | 0.9877 |
| LMFR-Net [51] | 2025 | 0.6626 | 0.9840 | 0.7971 | 0.9628 | 0.7350 | 0.9879 |
| FSE-Mamba [24] | 2025 | 0.6287 | 0.9802 | 0.7720 | 0.9597 | 0.6872 | **0.9897** |
| Rolling-Unet [47] | 2025 | 0 6798 | 0 9860 | 0 8094 | 0 9645 | 0 7579 | 0 9873 |
| MDF-Net [52] | 2026 | 0.6708 | 0.9735 | 0.8030 | 0.9607 | 0.8057 | 0.9778 |
| KM-UNet [53] | 2026 | 0.6812 | 0.9861 | 0.8104 | 0.9648 | 0.7579 | 0.9876 |
| LGFVM-UNet [48] | 2026 | 0.6312 | 0.9796 | 0.7739 | 0.9590 | 0.7055 | 0.9870 |
| Our | 2026 | **0.7023** | **0.9866** | **0.8251** | 0.9642 | **0.8484** | 0.9770 |

Table 5. Analysis of ablation experiment results for the model across three datasets

| Dataset | Methods | IOU | AUC | F1 | ACC | SE | SP |
|---|---|---|---|---|---|---|---|
| DRIVE | Baseline | 0.6223 | 0.9589 | 0.7672 | 0.9463 | 0.6948 | 0.9830 |
| | Baseline+PS-VSS | 0.7036 | 0.9801 | 0.8260 | 0.9570 | 0.8020 | 0.9796 |
| | Baseline+PS-VSS +SFCAM | 0.7076 | 0.9808 | 0.8288 | 0.9561 | 0.8356 | 0.9736 |
| STARE | Baseline | 0.6805 | 0.9761 | 0.8099 | 0.9541 | 0.7679 | 0.9813 |
| | Baseline+PS-VSS | 0.6831 | 0.9823 | 0.8117 | 0.9613 | 0.8069 | 0.9791 |
| | Baseline+PS-VSS +SFCAM | 0.7069 | 0.9840 | 0.8282 | 0.9634 | 0.8314 | 0.9791 |
| CHASE_DB1 | Baseline | 0.6568 | 0.9832 | 0.7929 | 0.9620 | 0.7330 | 0.9872 |
| | Baseline+PS-VSS | 0.6853 | 0.9869 | 0.8133 | 0.9653 | 0.7615 | 0.9877 |
| | Baseline+PS-VSS +SFCAM | 0.7023 | 0.9866 | 0.8251 | 0.9642 | 0.8484 | 0.9770 |

4 .3.3. Ablation Study

To assess the contributions of each module within our study, we conducted ablation experiments on three datasets, experimental results are presented in Table 5. Our model is fundamentally based on the UNet architecture, we maintained consistent all parameters and training settings for experimental validation.

Experimental results indicated that the proposed PS-VSS and SFCAM modules effec- tively improve retinal vessel segmentation performance. While the UNet model possesses basic vessel segmentation capabilities on three datasets, traditional convolutional opera- tions primarily focus on local neighbourhood features, there remain certain issues with missed detection in areas containing fine vessels, low-contrast vessels and complex topo- logical structures. The integration of the PS-VSS module resulted in improvements in IOU, AUC, F1, ACC and SE across all three datasets. For example, on the DRIVE dataset, IOU increased from 0.6223 to 0.7036, F1 rose from 0.7672 to 0.8260, and SE from 0.6948 to 0.8020. These findings demonstrate that the proposed Polygon-scanning SS2D method, through its sequential modelling approach that expands from the cen- tre outwards, can effectively capture the long-range dependencies and spatial continuity of vascular structures. Compared to traditional horizontal and vertical cross-scanning methods, it is better suited to describing the curved, branched, and slender topological features of retinal vessels.

Following the incorporation of SFCAM and PS-VSS modules, the model's IOU, F1 and SE were further improved. The complete model achieved IOU values of 0.7076, 0.7069 and 0.7023, F1 scores of 0.8288, 0.8282 and 0.8251, and SE values of 0.8356, 0.8314 and 0.8484 on three datasets, respectively. For instance, on the CHASE_DB1 dataset, the SFCAM increased the SE by 9.45% compared to the Baseline plus PS- VSS, indicating that this module significantly enhances the model's recall capability for vascular regions, particularly for fine vessels and areas with blurred edges. This performance improvement is linked to the collaborative modelling mechanism of spatial and frequency domain features within SFCAM.

It is noting that the SP of our model shows a slight decrease on some datasets. This phenomenon can be attributed to the model's improved sensitivity to vascular structures, which enables the successful detection of fine vessels and low-contrast vessels, albeit with a potential increase in misclassifications within background regions. However, improve- ments in SE and F1 typically better reflect the model's ability to comprehensively detect vascular structures. A comprehensive analysis of the IOU, AUC, F1, and SE results reveals that PS-VSS primarily enhances the model's global structural modelling capa-bilities, whilst SFCAM further strengthens the complementary representation of spatial details and frequency-domain information. The integration of these components effec- tively addresses the challenges of inadequate modeling of topological structures and in- sufficient utilization of detailed information during the transmission of complex features, thereby yielding more accurate, continuous, and detail-rich retinal vascular segmentation outcomes.

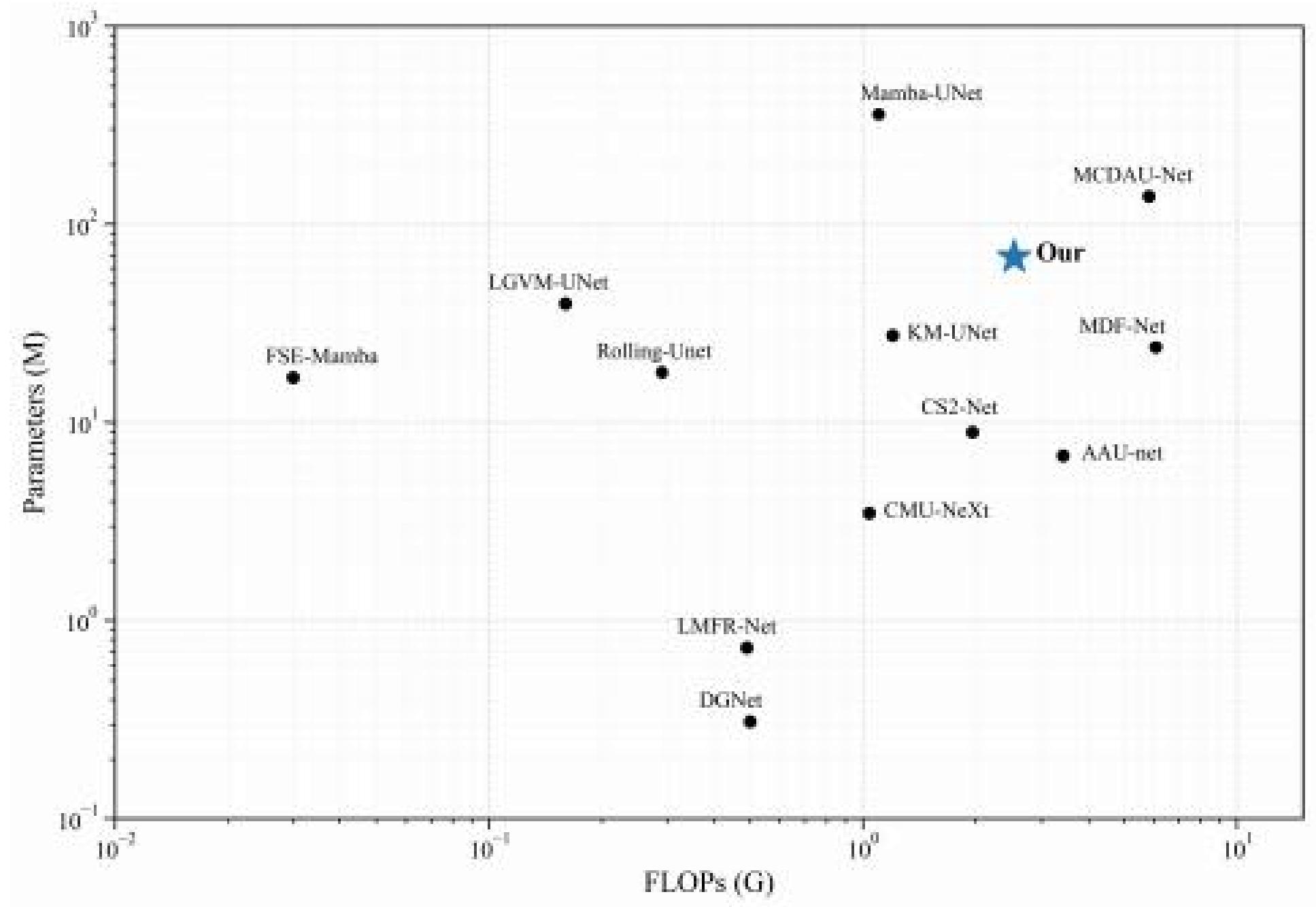

Figure 10. Parameters and FLOPs across different models.

4 .4 . Model complexity

The complexity comparison results indicate that our method involves 2.53 G FLOPs and a parameter count of 68.32 M, as shown in Fig. 10. Although the parameter count of our method is higher than some lightweight models, it remains significantly lower than some high computational complexity methods, such as MCDAU-Net [54] and MDF- Net [52]. Specifically, our method achieves a reduction in FLOPs by approximately 56.5% and 58.3% compared to MCDAU-Net[54] and MDF-Net[52], respectively. Furthermore, in comparison to Mamba-UNet[18], our method reduces the parameter count by approxi- mately 80.9%, demonstrating a more optimal balance between global modeling capability and parameter size. The increase in computational complexity in our method is primar- ily attributed to the integration of PS-VSS and SFCAM. PS-VSS enhances the modeling of long-range dependencies and complex topological structures in retinal vessels through polygonal scanning of SS2D, while SFCAM improves the representation of microvessels, edge contours, and local details through the synergistic fusion of spatial and frequency- domain features. Thus, our method does not solely aim to minimize computational cost but rather seeks to enhance the model's representation of vascular structure awareness within an acceptable range of complexity. A comprehensive analysis of experimental performance and complexity results demonstrates that our method achieves an optimal performance and complexity trade-off in the retinal vessel segmentation.

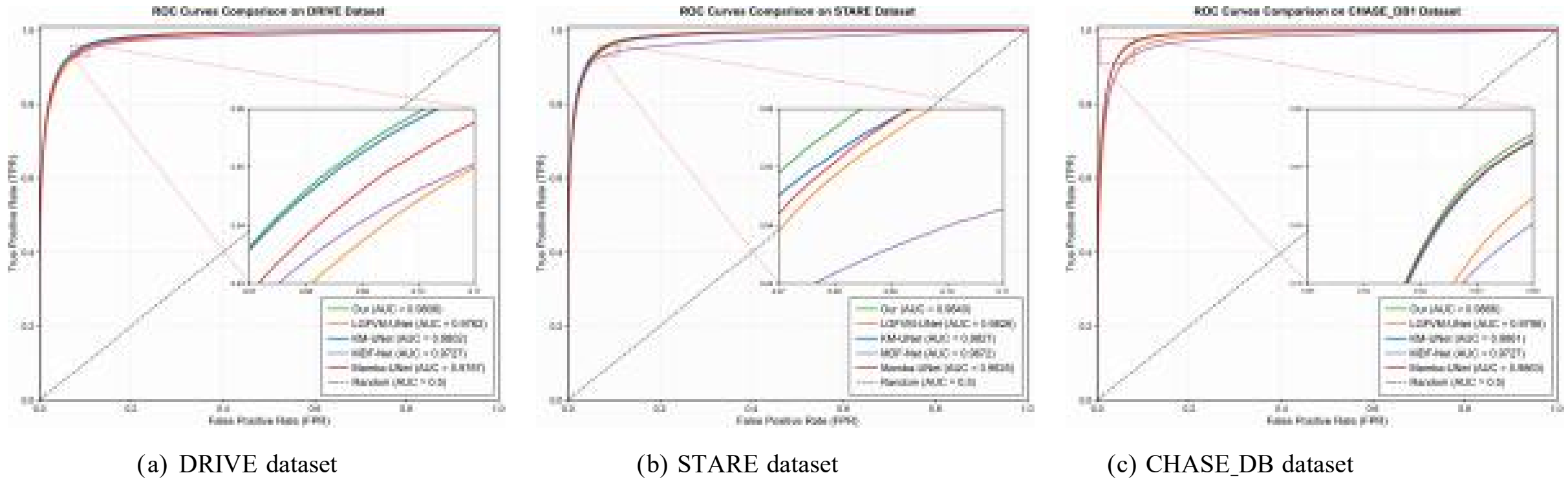

(a) DRIVE dataset (b) STARE dataset (c) CHASE_DB dataset

Figure 11. The Receiver Operating Characteristic (ROC) curves of different competitive models evaluated on three clinical benchmarking datasets.

### 4 .5. Visual analytics of the ROC curves

Fig 11 exhibited the ROC curves on the three datasets for five networks, including LGFVM-UNet [48], KM_UNet [53], MDF-Net [52], Mamba_Unet [18] and our model. The ROC curve illustrates the dynamic relationship between the true positive rate and the false positive rate, providing an objective measure of the model's proficiency in differenti- ating between retinal vessels and background. Our model obtained AUC values of 0.9808, 0.9840, and 0.9866 across the three datasets, representing the highest values among the compared methods. This indicates that our model exhibits robust competitive perfor- mance across three public datasets, underscoring its superior suitability for retinal vessel segmentation.

### 4 . 6. Evaluate polygon scanning mamba

To verify the proposed polygon scanning method's progressiveness and effectiveness, a comparative experiment was performed. As shown in Table 6, we validated the UNet, the traditional scanning method Mamba-UNet[18], and the proposed Triangle, Quadrilateral, Pentagon, Hexagon, and Octagon across three datasets.

On the DRIVE dataset, the Pentagon scanning method achieved the highest value with an IOU of 0.7036, F1 of 0.8260, ACC of 0.9570, and SE of 0.8020. Specifically, the F1 score of the Pentagon scanning method surpassed those of UNet and Mamba- UNet [18], while its SE value far exceeded the 0.6948 and 0.7847 values of UNet and Mamba-UNet [18]. This indicates that the Pentagon scanning method offers substan- tial advantages in capturing small microvessels and edge areas, thereby considerably reducing the rate of missed detections. On the STARE dataset, the Hexagon scanning method achieved the highest values in the four indicators of IoU, F1, ACC, and SE. The STARE dataset is known for its large differences in image contrast and interference from lesions. Hexagon's comprehensive leadership indicates that higher-order hexagonal scanning paths can provide richer neighborhood contextual information, which helps dis- tinguish vessels from lesion areas in complex backgrounds. On the dataset CHASE_DB1, Pentagon achieved the highest values in IoU, F1, ACC, and AUC, indicating that Pen- tagon has the best overall performance. In summary, the Pentagon scanning method achieved the highest values in multiple key indicators on the DRIVE and CHASE_DB1 datasets. Although the Pentagon scanning method has slightly weaker indicators than the Hexagon scanning method on the STARE dataset, it still has strong competitiveness on this dataset. Following a comprehensive evaluation of the three datasets, the Pentagon scanning method is the best choice for retinal vessel segmentation.

Table 6. Experimental evaluation of different scanning strategies.

| Dataset | Shape | IOU | AUC | F1 | ACC | SE | SP |
|---|---|---|---|---|---|---|---|
| DRIVE | UNet | 0.6223 | 0.9589 | 0.7672 | 0.9463 | 0.6948 | 0.9830 |
| | Mamba-UNet | 0.6937 | 0.9786 | 0.8191 | 0.9559 | 0.7847 | 0.9808 |
| | Triangle | 0.6995 | 0.9808 | 0.8232 | 0.9569 | 0.7889 | 0.9814 |
| | Quadrilateral | 0.7021 | 0.9796 | 0.8250 | 0.9570 | 0.7966 | 0.9804 |
| | Pentagon | 0.7036 | 0.9801 | 0.8260 | 0.9570 | 0.8020 | 0.9796 |
| | Hexagon | 0.7023 | 0.9801 | 0.8251 | 0.9570 | 0.7967 | 0.9804 |
| | Octagon | 0.7024 | 0.9803 | 0.8252 | 0.9569 | 0.8001 | 0.9797 |
| STARE | UNet | 0.6805 | 0.9761 | 0.8099 | 0.9541 | 0.7679 | 0.9830 |
| | Mamba-UNet | 0.6715 | 0.9825 | 0.8035 | 0.9617 | 0.7384 | 0.9881 |
| | Triangle | 0.6745 | 0.9828 | 0.8056 | 0.9616 | 0.7507 | 0.9866 |
| | Quadrilateral | 0.6815 | 0.9825 | 0.8106 | 0.9630 | 0.7465 | 0.9888 |
| | Pentagon | 0.6831 | 0.9823 | 0.8117 | 0.9613 | 0.8069 | 0.9791 |
| | Hexagon | 0.6978 | 0.9841 | 0.8220 | 0.9647 | 0.7692 | 0.9879 |
| | Octagon | 0.6903 | 0.9825 | 0.8168 | 0.9629 | 0.8011 | 0.9815 |
| CHASE_DB1 | UNet | 0.6568 | 0.9832 | 0.7929 | 0.9620 | 0.7330 | 0.9872 |
| | Mamba-UNet | 0.6788 | 0.9863 | 0.8087 | 0.9645 | 0.7542 | 0.9877 |
| | Triangle | 0.6748 | 0.9859 | 0.8058 | 0.9642 | 0.7471 | 0.9882 |
| | Quadrilateral | 0.6831 | 0.9085 | 0.8117 | 0.9650 | 0.7598 | 0.9876 |
| | Pentagon | 0.6854 | 0.9869 | 0.8133 | 0.9653 | 0.7615 | 0.9877 |
| | Hexagon | 0.6832 | 0.9865 | 0.8118 | 0.9649 | 0.7628 | 0.9871 |
| | Octagon | 0.6849 | 0.9862 | 0.8130 | 0.9647 | 0.7714 | 0.9806 |

## 5. Discussion

This article presents an algorithm for retinal vessel segmentation, leveraging PS-VSS and SFCAM module, which has several specific advantages and characteristics. (I) A unique polygon scanning mamba is proposed to effectively extract the global receptive field while maintaining linear complexity. However, traditional mamba scanning oper- ations from both horizontal and vertical directions may result in the absence of small vessels. (II) By deeply fusing frequency attention and spatial attention, our model fully exploits frequency attention to extract effective features, and leverages spatial attention to focus positional and structural features, effectively extracting complex topological struc- ture features, texture features, and edge features of the fundus vessels. In contrast, most methods rely on single attention, which may lead to the loss of partial fundus vascular information. (III) The merits of the CNN, PS-VSS and SFCAM are closely combined to create an effective segmentation framework for retinal vessel detection. (IV) Our scheme can effectively preserve the integrity of fine vessels while maintaining the topological features of the retinal vessels.

Our model was tested on three datasets. Experimental findings indicated that our scheme has the best performance in retinal vessel detection than the comparative methods [18, 24, 34, 48, 50, 51, 52, 53], as evidenced by both visual inspection and quantitative analysis. The key reason is that most methods are modifications of the traditional UNet model, which are not effective at capturing long-range dependencies. While Mamba- UNet [18], FSE-Mamba [24] and LGFVM-UNet[48] are capable of solving this problem efficiently, but they does not perform well in capturing the intricate topological features of small vessels. In contrast, our model integrates the PS-VSS module and SFCAM module into the UNet model, which can effectively segment small retinal vessels.

Despite the promising performance of our model in retinal vessel segmentation, there are also some limitations. (I) Our model employs a hybrid CNN-Mamba fusion strategy, which may lead to increased computational complexity and parameter count. (II) Al- though datasets are being expanded by data augmentation techniques, the generalization capability of these models is still limited by the relatively few images available for train- ing and testing. Despite these limitations, our model still delivers the best outcomes for segmenting retinal vessels.

## 6. Conclusion

This study introduces a novel algorithm using polygon scanning mamba and space-frequency collaborative attention for retinal vessel segmentation. Initially, an innovative scanning strategy based on Mamba architecture is presented to identify complex struc- tural features, thereby managing long-distance dependencies while maintaining linear complexity. Subsequently, an improved space-frequency collaborative attention is devel- oped to extract vessel details from spatial and frequency domain, significantly decreasing the loss of information from small vessels. Our model was tested on public datasets, and the experimental outcomes demonstrated that our scheme has the best performance in terms of small vessel detection. In the future, we will focus on improving SP while main- taining SE. By leveraging the structural characteristics of retinal vessels, we will introduce traditional pattern recognition methods to suppress noise. Concurrently, more advanced deep learning models will be designed to effectively segment small retinal vessels.

## CRediT authorship contribution statement

**Yuanyuan Peng:** Writing – review & editing, Supervision, Funding acquisition. **Wen Li:** Writing – original draft, Visualization, Software, Methodology.

## Declaration of competing interest

The authors declare that they have no known competing financial interests or personal relationships that could have appeared to influence the work reported in this paper.

## Acknowledgements

This article was supported by Jiangxi Provincial Natural Science Foundation (nos. 20242BAB25057).

## CONFLICT OF INTEREST STATEMENT

The authors declare no conflicts of interest.

## Data availability statement

These data were derived from the following resources available in the public domain:
DRIVE: https://drive.grand-challenge.org/;
STARE: http://www.ces.clemson.edu/ahoover/stare/;
CHASED_DB1: https://blogs.kingston.ac.uk/retinal/chasedb1/.

## ORCID

**Yuanyuan Peng:** https://orcid.org/0000-0003-2154-9759.
**Wen Li:**https://orcid.org/0009-0005-8559-7890.